\documentclass[lettersize,journal]{IEEEtran}
\usepackage{amsmath,amsfonts}
\usepackage{algorithmic}
\usepackage{array}
\usepackage[caption=false,font=normalsize,labelfont=sf,textfont=sf]{subfig}
\usepackage{textcomp}
\usepackage{stfloats}
\usepackage{url}
\usepackage{verbatim}
\usepackage{graphicx}
\def\BibTeX{{\rm B\kern-.05em{\sc i\kern-.025em b}\kern-.08em
    T\kern-.1667em\lower.7ex\hbox{E}\kern-.125emX}}
\usepackage{balance}
\usepackage{booktabs}

\renewcommand{\baselinestretch}{1.02}
\begin{document}
\title{Progressive Image Restoration via Text-Conditioned Video Generation
}

\author{
	\IEEEauthorblockN{
		Peng Kang\IEEEauthorrefmark{1}\textsuperscript{1},
		Xijun Wang\IEEEauthorrefmark{2}\textsuperscript{1},
		Yu Yuan\IEEEauthorrefmark{2} \\
	}
	\IEEEauthorblockA{
		\IEEEauthorrefmark{1}Department of Computer Science, University of Illinois Springfield, Springfield, IL, USA \\
		\IEEEauthorrefmark{2}School of Electrical and Computer Engineering, Purdue University, West Lafayette, IN, USA
	}
	\thanks{\textsuperscript{1}Both authors contributed equally to this work. Peng Kang is the corresponding author. Email: pkang8@uis.edu}
}

\maketitle

\begin{abstract}
	Recent text-to-video models have demonstrated strong temporal generation capabilities, yet their potential for image restoration remains underexplored. In this work, we repurpose CogVideo for progressive visual restoration tasks by fine-tuning it to generate restoration trajectories rather than natural video motion. Specifically, we construct synthetic datasets for super-resolution, deblurring, and low-light enhancement, where each sample depicts a gradual transition from degraded to clean frames. Two prompting strategies are compared: a uniform text prompt shared across all samples, and a scene-specific prompting scheme generated via LLaVA multi-modal LLM and refined with ChatGPT. Our fine-tuned model learns to associate temporal progression with restoration quality, producing sequences that improve perceptual metrics such as PSNR, SSIM, and LPIPS across frames. Extensive experiments show that CogVideo effectively restores spatial detail and illumination consistency while maintaining temporal coherence. Moreover, the model generalizes to real-world scenarios on the ReLoBlur dataset without additional training, demonstrating strong zero-shot robustness and interpretability through temporal restoration.
\end{abstract}


\section{Introduction}
\label{sec:intro}

Recent advances in text-to-video diffusion models~\cite{yuan2025identity, ruan2024enhancing, lee2024grid, yuan2025magictime, chen2024videocrafter2, wang2024recipe, yuan2024instructvideo} have enabled the generation of visually coherent and temporally consistent video content directly from textual prompts. Among these models, CogVideo~\cite{yang2024cogvideox} has demonstrated remarkable capability in synthesizing high-quality videos that capture realistic motion dynamics and scene details. While such models have primarily been used for creative video generation, their potential for visual restoration and enhancement tasks remains largely unexplored.

In this work, we propose a novel paradigm that leverages CogVideo as a generative bridge for visual enhancement tasks, including image super-resolution, deblurring, and low-light enhancement. Instead of treating these tasks as independent image-to-image translation problems, we view them through the lens of video progression learning: that is, generating a sequence of temporally ordered frames that evolve from degraded inputs (e.g., low-resolution, dark, or blurred) toward their enhanced versions. By fine-tuning CogVideo on specialized datasets that encode these degradations as temporal transitions, the model learns to capture the intrinsic relationships between visual degradation and recovery in a dynamic, context-aware manner.

Our key insight is that temporal consistency and motion priors learned by CogVideo can be exploited to improve spatial detail recovery and stability in single-frame restoration tasks. Traditional restoration methods~\cite{zamir2022restormer, kim2016accurate, tai2017memnet, liu2018non, lefkimmiatis2018universal}, including CNN- and transformer-based architectures, often focus on spatial fidelity without modeling progressive transitions. In contrast, our approach harnesses the temporal diffusion process of CogVideo to implicitly model the gradual transformation from degraded to clean states, yielding results that are both perceptually smooth and detail-preserving.

To evaluate this idea, we construct and fine-tune CogVideo on three newly formalized datasets:
\begin{itemize}
     \item Resolution Progression Dataset – videos simulating transitions from low to high resolution;
     \item Illumination Progression Dataset – videos progressing from dark to well-lit conditions; 
     \item Blur-to-Sharp Dataset – videos depicting gradual recovery from motion or defocus blur.
\end{itemize}

Comprehensive experiments show that our fine-tuned CogVideo not only produces visually appealing videos but also serves as a powerful generator for restoration tasks when sampled frame-by-frame. Furthermore, to assess the generalization of our approach, we apply the fine-tuned CogVideo model to the ReLoBlur dataset~\cite{li2023real}, a challenging real-world video deblurring benchmark. Despite not being explicitly trained on real-world motion blur, our model demonstrates superior visual restoration and temporal stability, producing realistic and coherent enhancement sequences.

In summary, the main contributions of this work are threefold:
\begin{itemize}
     \item We propose a novel framework that repurposes CogVideo for image enhancement tasks by modeling restoration as a video progression process.
     \item We construct and release three benchmark datasets representing progressive visual degradation-to-restoration transitions.
     \item We demonstrate that fine-tuned CogVideo achieves competitive performance in super-resolution, deblurring, and low-light enhancement, offering a new perspective on leveraging text-to-video diffusion models for image restoration.
 \end{itemize}

\section{Methodology}
\label{sec:methodology}
Building upon the motivation introduced earlier, our goal is to repurpose CogVideo, a large-scale text-to-video diffusion model, for image restoration and enhancement tasks. Rather than designing a new network architecture from scratch, we exploit the generative prior and temporal reasoning capabilities already embedded in CogVideo. By fine-tuning it on carefully constructed progression datasets, we guide the model to interpret “visual enhancement” as a gradual temporal transformation.

Our method proceeds in three stages. First, we formalize the restoration problem as a temporal video generation task, where a degraded image progressively evolves into its clean counterpart. Second, we construct specialized datasets for super-resolution, deblurring, and low-light correction, providing both uniform and scene-adaptive text conditions for comparison. Finally, we fine-tune CogVideo using a framework that conditions the diffusion process on textual descriptions of the restoration. The overall pipeline is designed to be modular and data-driven, allowing CogVideo to generalize across multiple degradation types.

\subsection{Problem Definition}

Given a degraded image $I_d$ (e.g., low-resolution, blurred, or low-light) and its corresponding clean target $I_c$, conventional restoration methods aim to learn a direct mapping
\begin{equation}
	f_\theta: I_d \rightarrow I_c,
\end{equation}
where $f_\theta$ is typically a CNN- or Transformer-based model optimized for pixel-level fidelity or perceptual similarity.

In this work, we reinterpret image restoration as a \textbf{temporal video generation} problem. Instead of predicting a single restored frame, we model the progressive enhancement process as a temporally ordered sequence
\begin{equation}
	\mathcal{V} = \{ I_1, I_2, \dots, I_T \},
\end{equation}
where $I_1 = I_d$ is the degraded input, $I_T = I_c$ is the restored target, and the intermediate frames $\{I_t\}_{t=2}^{T-1}$ represent gradual transitions from degraded to enhanced states.

Let $G_\phi$ denote the CogVideo generator parameterized by $\phi$. During inference, the model takes as input both a textual prompt $p$ that semantically describes the desired restoration process and the initial degraded frame $I_d$. The generation process can thus be formulated as
\begin{equation}
	\mathcal{V} = G_\phi(p, I_d),
\end{equation}
where $\mathcal{V}$ is a short video depicting a smooth enhancement trajectory from $I_d$ to $I_T$.
The final frame $I_T$ is extracted as the restored image, while preceding frames visualize interpretable intermediate restoration stages.

This formulation offers two key advantages. 
First, by leveraging the temporal diffusion dynamics of CogVideo, the model captures the gradual, continuous nature of visual enhancement. Second, the text prompt $p$ provides an additional conditioning signal that aligns the generation process. Through this joint conditioning, the model learns to couple textual guidance with spatiotemporal visual transformations, enabling a unified framework for super-resolution, deblurring, and low-light enhancement.

\begin{figure*}[htp]
	\centering
	\includegraphics[width=0.95\textwidth]{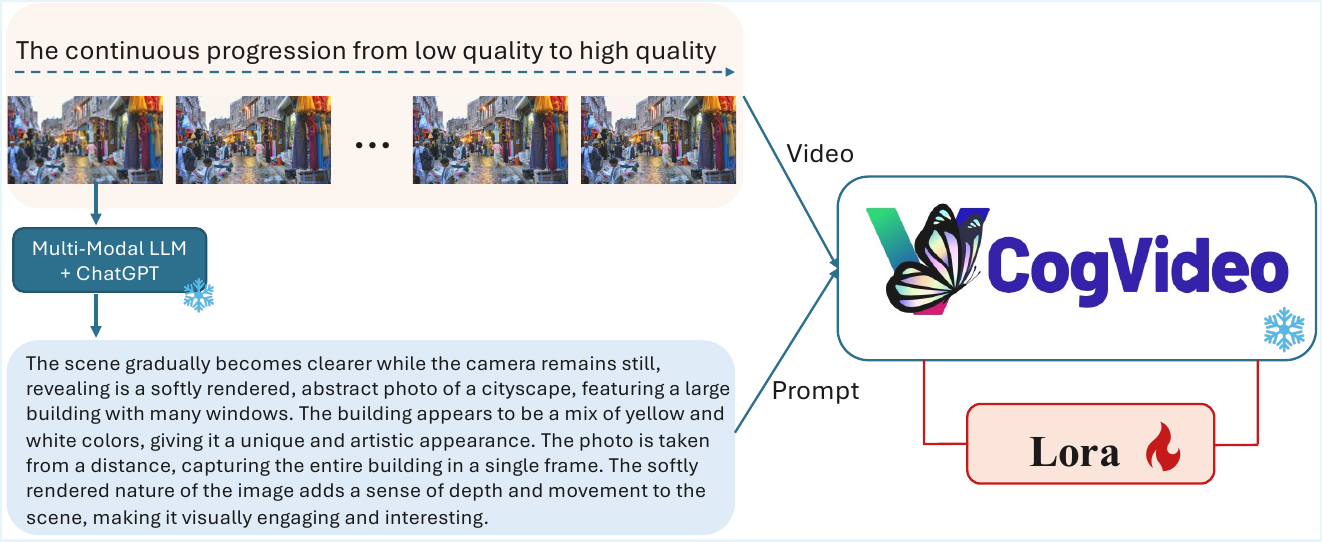}
	\caption{Illustration of the CogVideo fine-tuning process for image restoration. The model receives a video showing progressive visual enhancement (e.g., from low-resolution to super-resolution) and a corresponding textual prompt describing the scene and restoration dynamics. Using LoRA fine-tuning, CogVideo learns to align temporal visual improvements with text semantics, enabling it to generate restoration sequences conditioned on both the input frame and the prompt.}
	\label{fig:pipeline}
\end{figure*}

\subsection{Dataset Construction}

To enable CogVideo to learn visual restoration as a progressive video generation task, we construct three domain-specific datasets derived from the DIV2K high-resolution image corpus~\cite{Timofte_2017_CVPR_Workshops}. Each dataset contains short video clips, each composed of $T=9$ frames showing a smooth transition from a degraded state to a clean target. For every image in the DIV2K dataset, we generate two distinct videos using random degradation parameters to increase diversity. All generated videos are stored at $1360\times768$ resolution and 5~fps.

\vspace{0.5em}
\noindent\textbf{1) Resolution Progression Dataset.}
We simulate low-to-high resolution transitions by progressively upscaling bicubic-downsampled images. Each video is generated by resizing the image with scale factors $\{s_1, s_2, \ldots, s_9\}$ linearly increasing from a random minimum $s_1\in[0.05,0.25]$ to $s_9=1.0$. JPEG compression artifacts are added at lower resolutions to mimic real-world degradation. The corresponding Python implementation (simplified) is:
\begin{equation}
	I_t = \text{Resize}\big(\text{Resize}(I_{\text{HR}}, s_t), 1/s_t\big), \quad t=1,\ldots,9.
\end{equation}

\vspace{0.5em}
\noindent\textbf{2) Blur-to-Sharp Dataset.}
To emulate motion blur recovery, we apply directional motion kernels with gradually decreasing blur intensity. For each image, a random blur angle $\alpha\in[0,360)$ and a maximum kernel length $k_{\max}\in[40,200]$ are sampled. Each frame $I_t$ is obtained by applying a kernel of size $k_t = k_{\max}\!\cdot\!(1 - (t-1)/8)$. The last frame ($t=9$) is left unblurred.

\begin{equation}
	I_t = \text{MotionBlur}(I_{\text{HR}}, k_t, \alpha), \quad t=1,\ldots,9.
\end{equation}

\vspace{0.5em}
\noindent\textbf{3) Low-Light Progression Dataset.}
For low-light enhancement, we generate realistic darkening effects using exposure roll-off, white-balance drift, Gaussian noise, and mild blur. Each video depicts a gradual illumination increase from dark to bright:
\begin{equation}
	I_t = \mathcal{L}(I_{\text{HR}}, s_t, n_t),
\end{equation}
where $\mathcal{L}(\cdot)$ represents the low-light transformation parameterized by strength $s_t=1-(t-1)/8$ and noise level $n_t\!\in\![0.02,0.08]$. This formulation produces visually convincing nighttime conditions with sensor noise and color casts.

\vspace{0.5em}
\noindent\textbf{Data Organization.}
Each dataset directory contains the generated videos, along with two text files: \texttt{videos.txt} listing relative paths to all clips, and \texttt{prompt.txt} containing the corresponding textual prompts. For every image, two videos are generated with random degradation strengths, resulting in twice the number of samples as the source dataset.

\vspace{0.5em}
\noindent\textbf{Uniform vs.~Scene-Adaptive Prompts.}
We prepare two prompt versions for fine-tuning:
\begin{itemize}
	\item \textbf{Uniform Text Version:} All videos within the same task share a fixed prompt, e.g., \textit{``The image becomes sharper and higher in resolution. Nothing moves. Static image.''} for the resolution dataset.
	\item \textbf{Scene-Adaptive Text Version:} Each video uses a distinct prompt generated automatically by LLaVA \cite{Liu2023VisualIT} and refined with ChatGPT~\cite{radford2018improving, radford2019language, brown2020language, achiam2023gpt}, incorporating semantic details from the scene while maintaining the same restoration theme (e.g., ``A night street gradually brightens under lamplight.'' or ``A blurred portrait becomes focused.'').
\end{itemize}

\begin{figure*}
	\centering
	\includegraphics[width=0.32\textwidth]{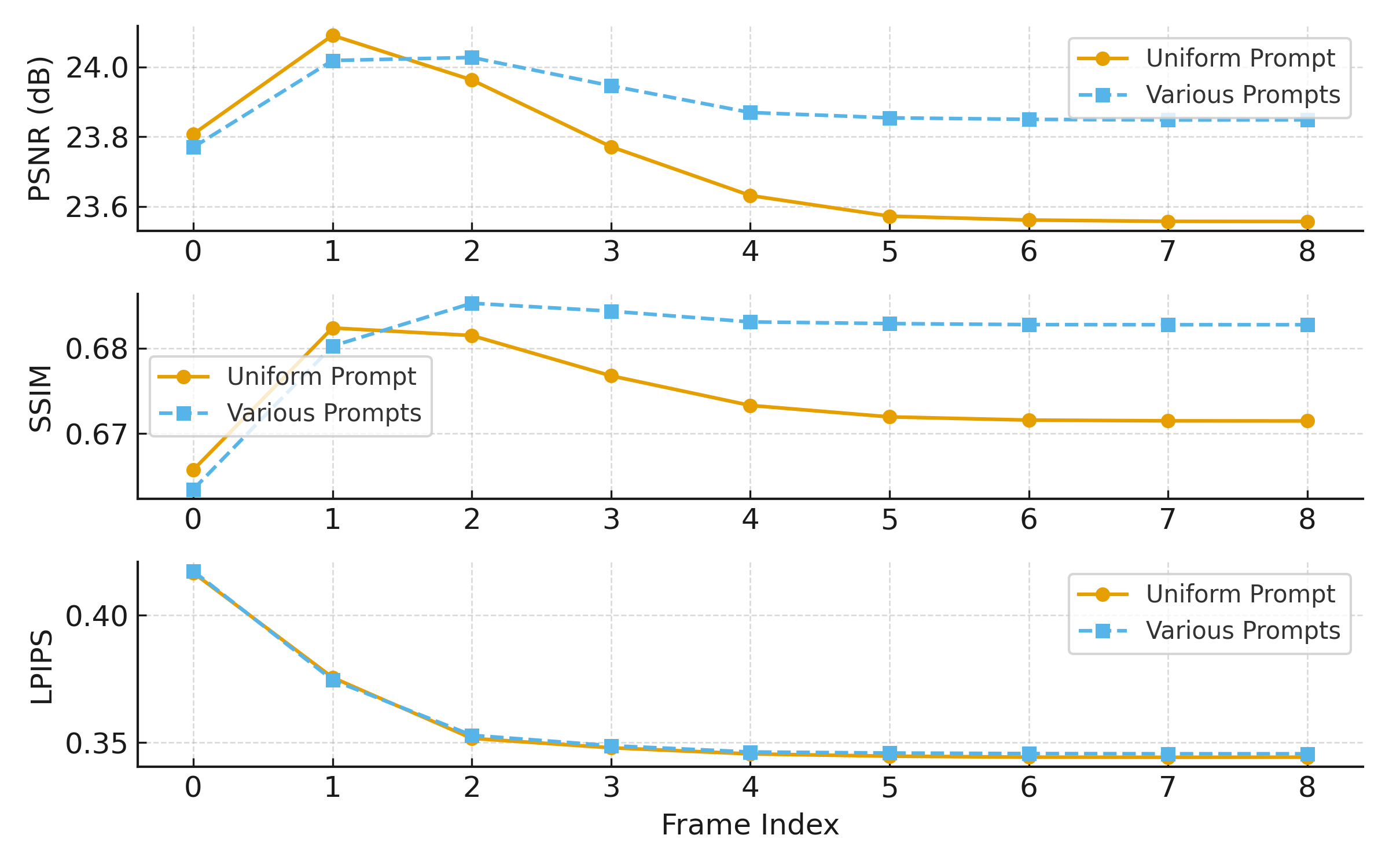}
	\includegraphics[width=0.32\textwidth]{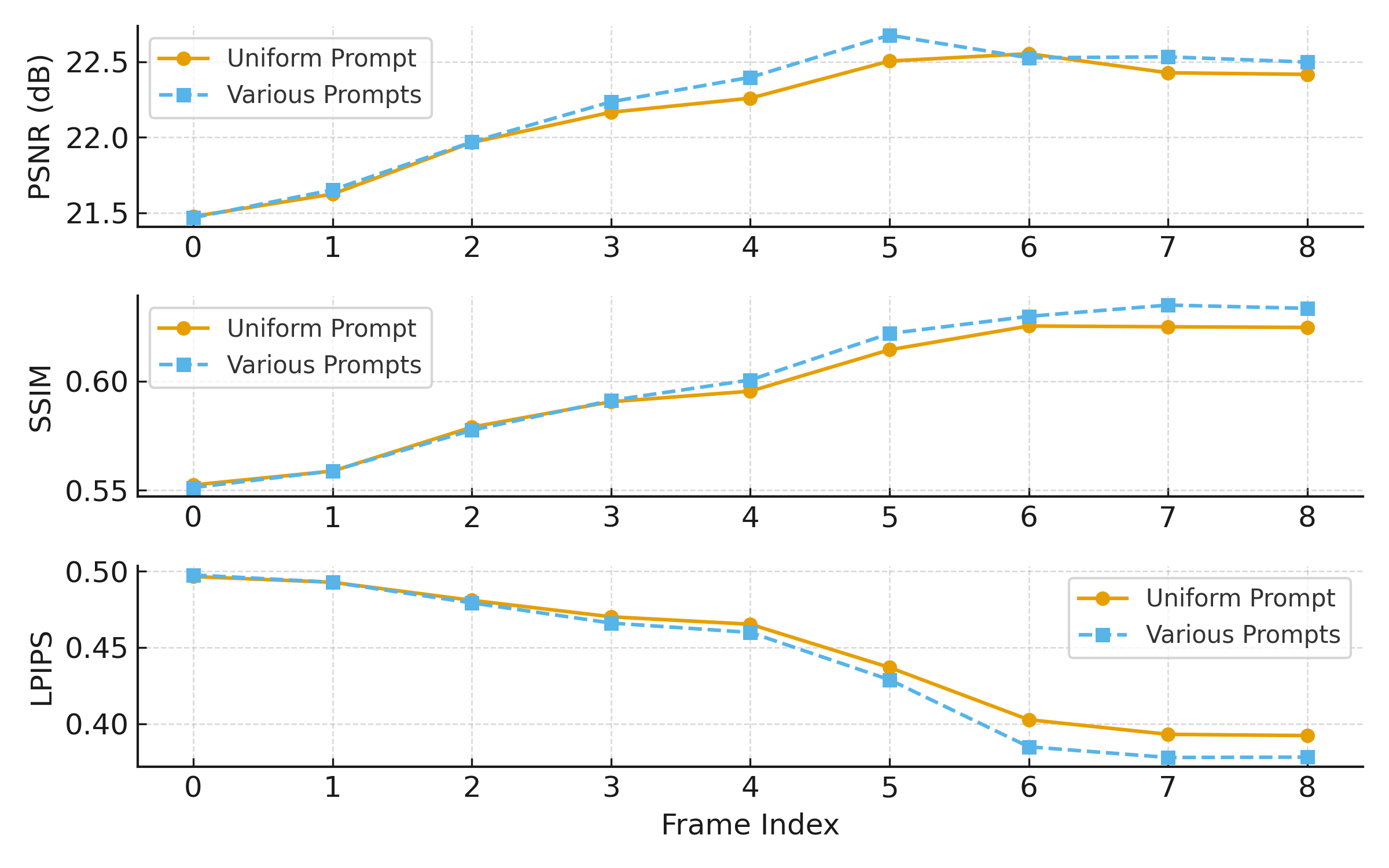}
	\includegraphics[width=0.32\textwidth]{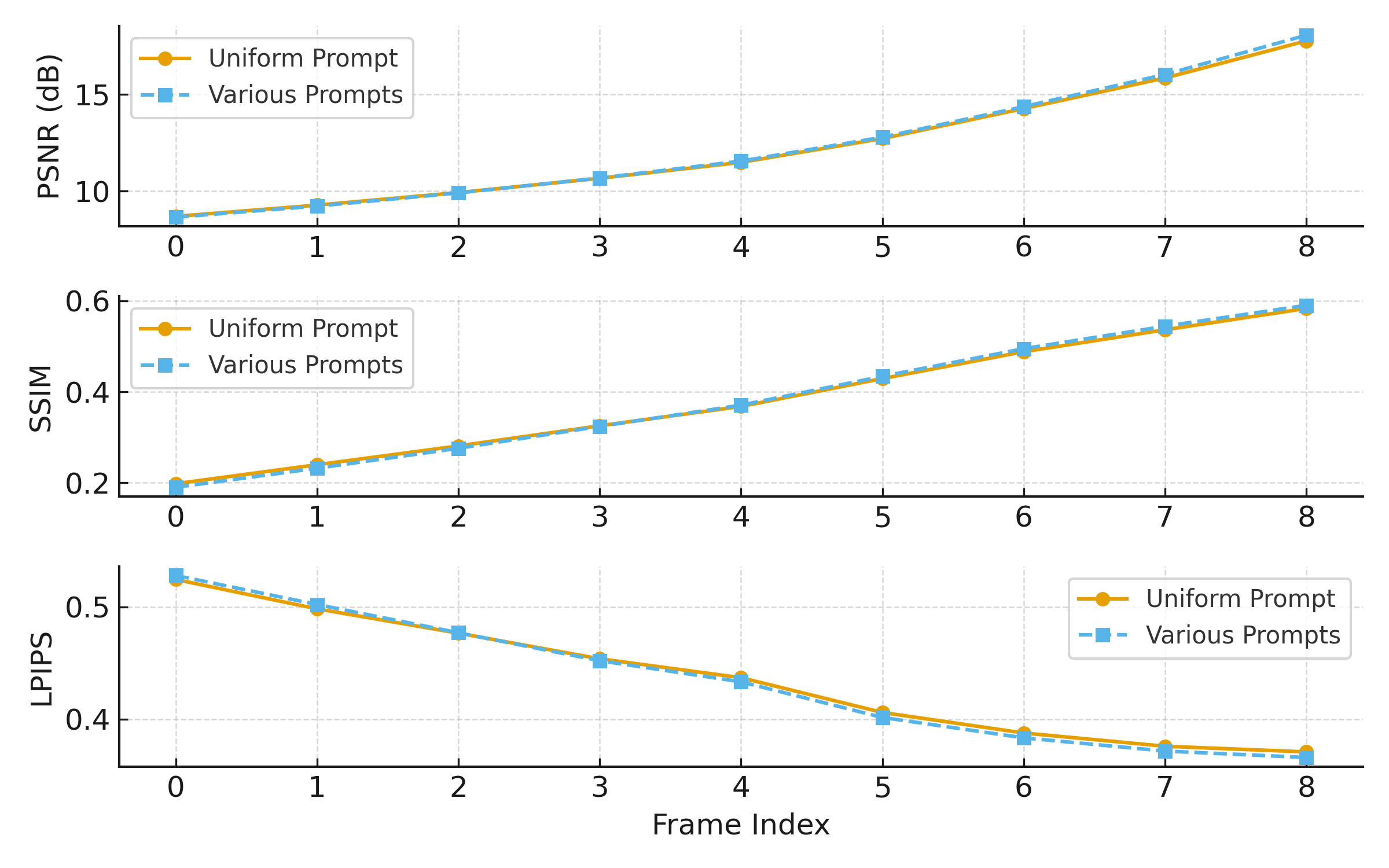}
	\caption{
		Frame-wise restoration performance across three enhancement tasks: (a) super-resolution, (b) deblurring, and (c) low-light enhancement. For each task, PSNR and SSIM generally increase while LPIPS decreases as frames progress, confirming that the model learns temporal enhancement dynamics. The various-prompt version consistently provides slightly higher perceptual quality and smoother progression, highlighting the benefit of scene-aware textual conditioning.
	}
	\label{fig:all_quantitative}
\end{figure*}

\subsection{Fine-Tuning and Inference}

We fine-tune the pretrained CogVideo model to learn progressive restoration from degraded to clean images. The overall pipeline is illustrated in Fig.~\ref{fig:pipeline}. Given a textual prompt $p$ describing the enhancement process and an initial degraded image $I_d$, the model generates a video sequence 
$\mathcal{V} = \{ I_1, I_2, \dots, I_T \}$ that depicts a smooth visual transition toward restoration. The last frame $I_T$ serves as the final restored output. 

We employ lightweight LoRA fine-tuning to adapt CogVideo efficiently while keeping most parameters frozen. 

\vspace{0.5em}
\noindent\textbf{Inference.}
At inference time, CogVideo takes a degraded input image $I_d$ and a prompt $p$ (either uniform or scene-adaptive) and generates a video $\mathcal{V} = G_\phi(p, I_d)$. The resulting frames $\{ I_t \}$ visualize the restoration trajectory, and the last frame $I_T$ is used as the enhanced output image. This process enables unified inference for super-resolution, deblurring, and low-light enhancement within a single generative framework.

\section{Experiments}
\label{sec:experiments}

We conduct a series of experiments to evaluate the effectiveness of our fine-tuned CogVideo framework on three synthetic datasets (super-resolution, deblurring, and low-light enhancement) and one real-world dataset for motion blur restoration. We report both quantitative metrics and qualitative visual comparisons. All experiments are performed on one NVIDIA H100 GPU using the LoRA fine-tuned CogVideoX-5B-I2V model. 

\begin{table*}[t]
	\centering
	\caption{
		Frame-wise LPIPS values on the ReLoBlur test set. 
		Lower LPIPS indicates better performance.
	}
	\label{tab:real_lpips}
	\begin{tabular}{cccccccccc}
		\toprule
		Frame & 1 & 2 & 3 & 4 & 5 & 6 & 7 & 8 & 9 \\
		\midrule
		LPIPS & 0.335 & 0.330 & 0.323 & 0.319 & 0.316 & 0.318 & 0.316 & 0.317 & 0.317 \\
		\bottomrule
	\end{tabular}
\end{table*}

\begin{figure*}[t]
	\centering
	\setlength{\tabcolsep}{1pt}
	\renewcommand{\arraystretch}{0.5}
	\begin{tabular}{ccccc}
		\includegraphics[width=0.19\textwidth]{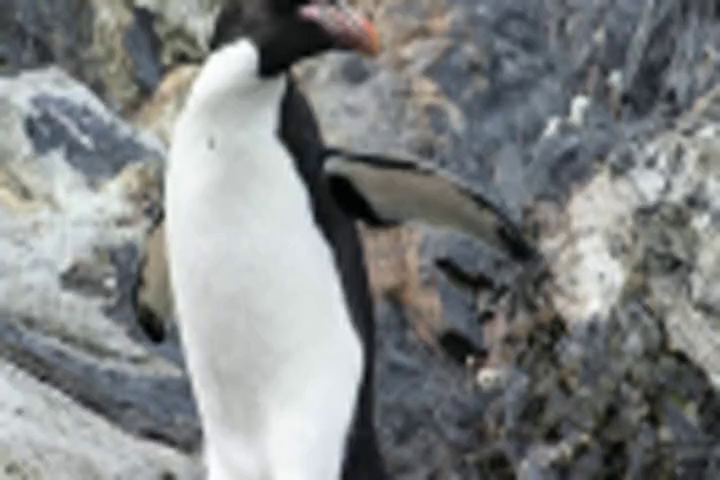} &
		\includegraphics[width=0.19\textwidth]{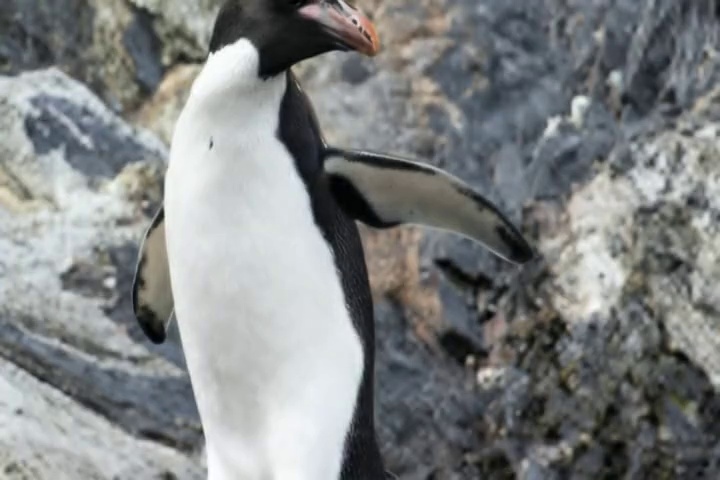} &
		\includegraphics[width=0.19\textwidth]{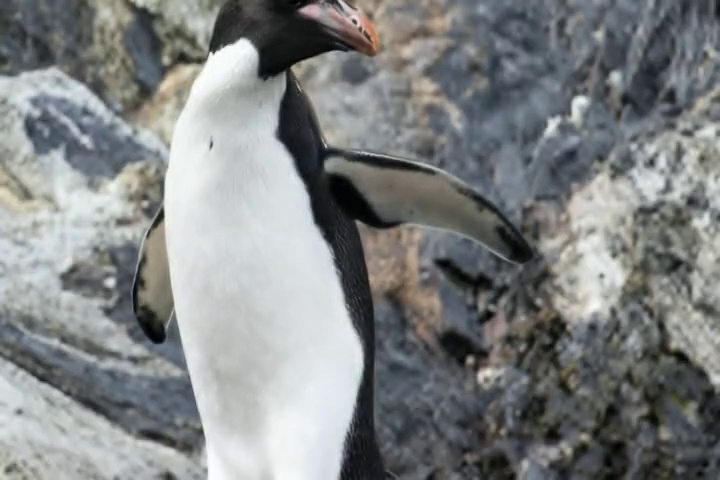} &
		\includegraphics[width=0.19\textwidth]{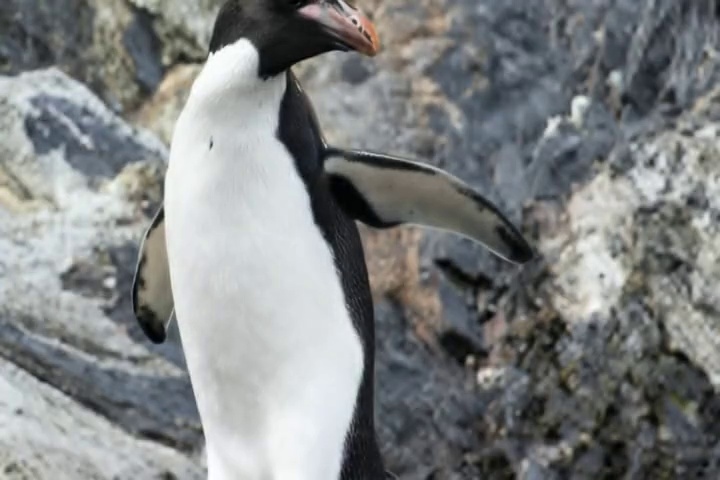} &
		\includegraphics[width=0.19\textwidth]{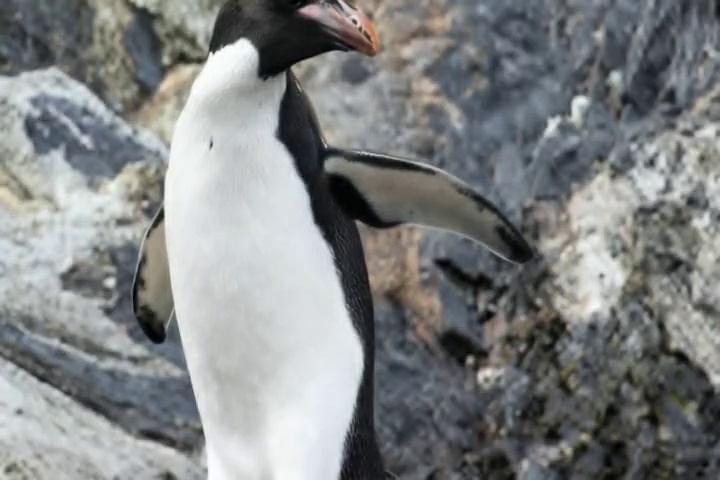} \\
		\multicolumn{5}{c}{\small (a) Super-Resolution Progression} \\[0.6em]
		\includegraphics[width=0.19\textwidth]{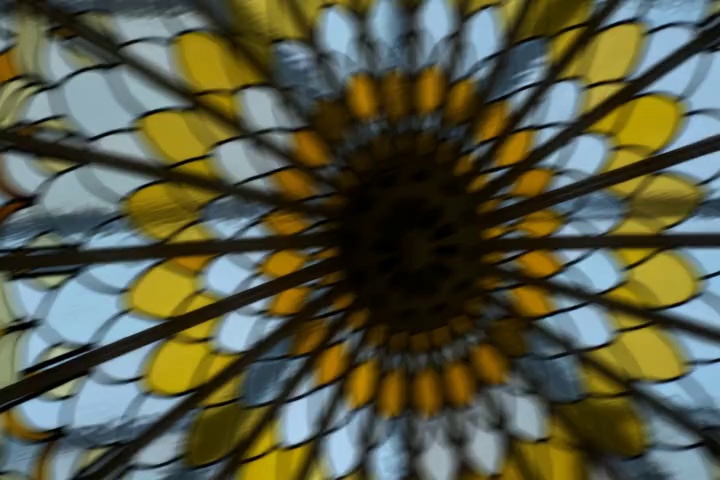} &
		\includegraphics[width=0.19\textwidth]{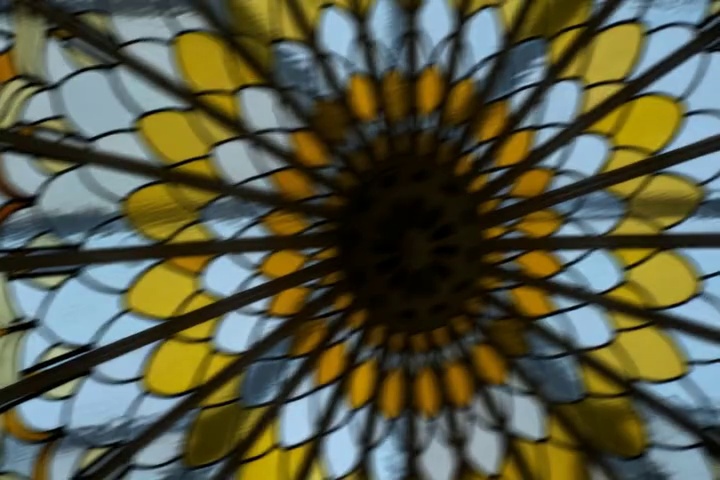} &
		\includegraphics[width=0.19\textwidth]{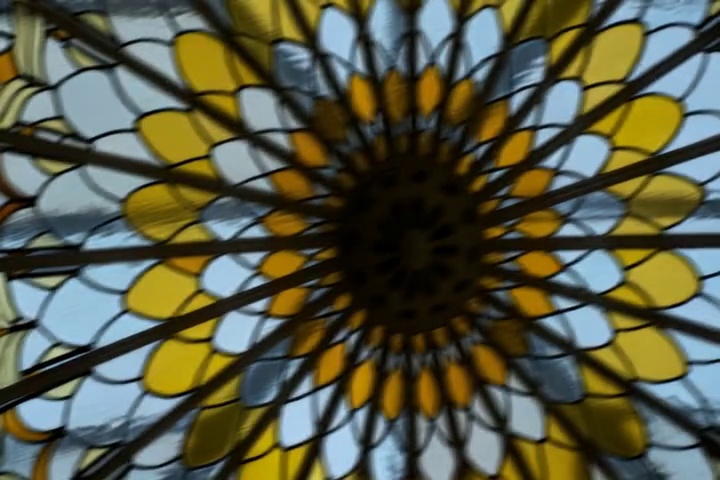} &
		\includegraphics[width=0.19\textwidth]{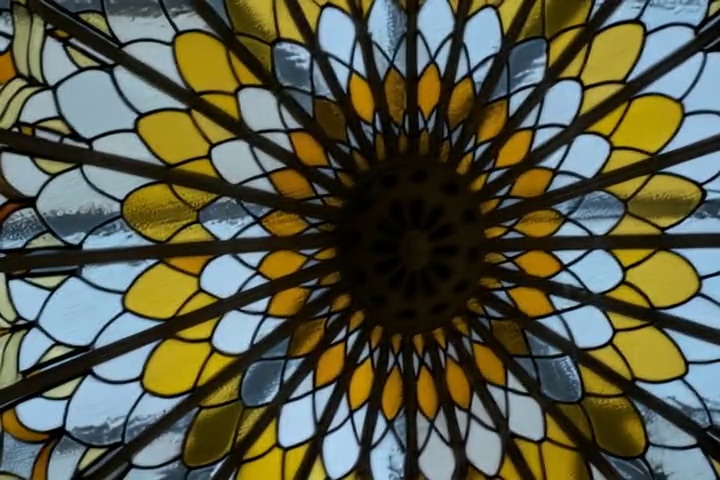} &
		\includegraphics[width=0.19\textwidth]{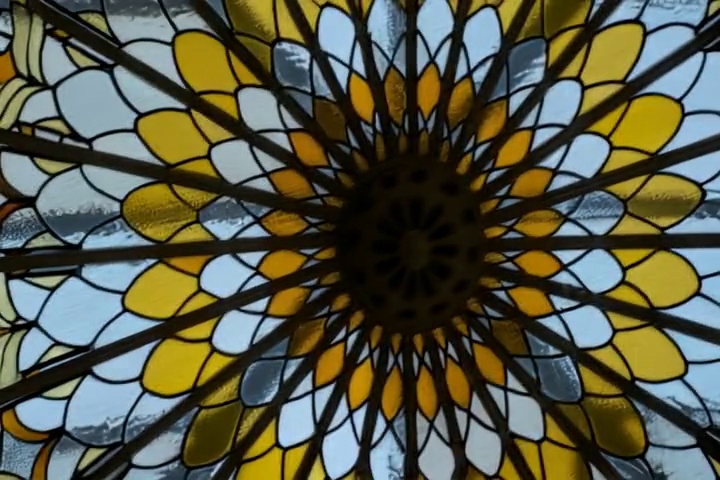} \\
		\multicolumn{5}{c}{\small (b) Deblurring Progression} \\[0.6em]
		\includegraphics[width=0.19\textwidth]{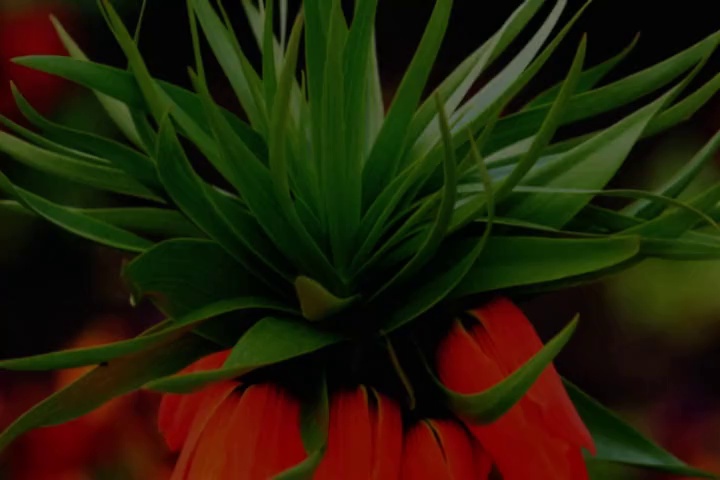} &
		\includegraphics[width=0.19\textwidth]{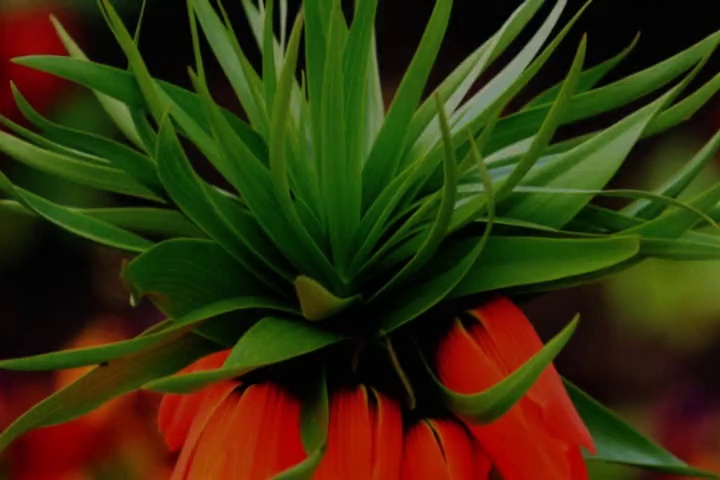} &
		\includegraphics[width=0.19\textwidth]{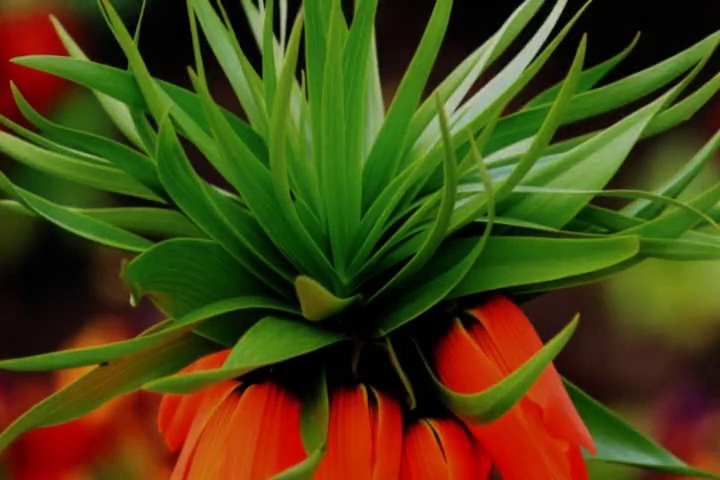} &
		\includegraphics[width=0.19\textwidth]{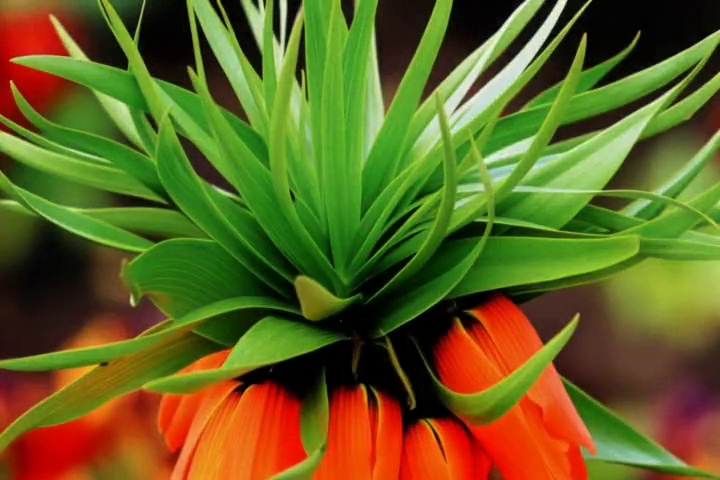} &
		\includegraphics[width=0.19\textwidth]{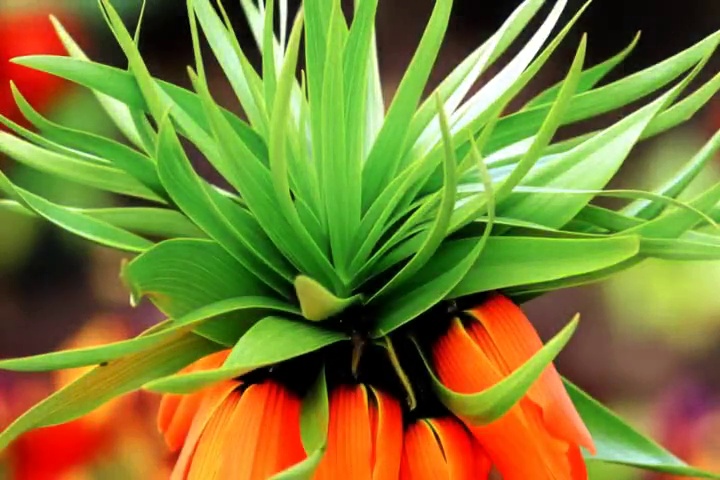} \\
		\multicolumn{5}{c}{\small (c) Low-Light Enhancement Progression} \\[0.2em]
	\end{tabular}
	\caption{
		Qualitative restoration results across three enhancement tasks using the various-prompt fine-tuned CogVideo.
		Each row visualizes frames 1, 3, 5, 7, and 9 to illustrate the temporal restoration trajectory. The model demonstrates smooth progression, gradually recovering fine details, edges, and illumination while preserving global structure and perceptual consistency.
	}
	\label{fig:qualitative_all}
\end{figure*}

\subsection{Quantitative Analysis}

To comprehensively evaluate the restoration dynamics across different degradation types, we summarize the frame-wise quantitative trends of the fine-tuned CogVideo on the super-resolution, deblurring, and low-light enhancement datasets. Figure~\ref{fig:all_quantitative} visualizes PSNR, SSIM, and LPIPS progression curves for both the uniform- and scene-adaptive various-prompt configurations across all three tasks.

Across all tasks, the fine-tuned CogVideo exhibits a clear temporal enhancement trend: initial frames correspond to heavily degraded conditions, while later frames progressively restore spatial details and perceptual fidelity. In the super-resolution setting, performance improves rapidly during early frames and then stabilizes, indicating fast convergence toward the clean state. For deblurring, the progression is smoother and more gradual, suggesting that temporal modeling helps capture the evolution of motion sharpness. In low-light enhancement, the improvement is monotonic across all frames, reflecting the model's ability to simulate illumination recovery over time.

Comparing prompt strategies, the various-prompt version consistently achieves higher PSNR and SSIM and lower LPIPS across all three datasets, demonstrating better generalization and semantic-text alignment. This confirms that integrating diverse, scene-specific prompts during fine-tuning encourages the model to more effectively couple textual context with visual enhancement, leading to both quantitative and perceptual gains.

\subsection{Qualitative Results}

To further demonstrate the restoration capability and temporal coherence of the proposed framework, we visualize frame-wise qualitative results from the various-prompt fine-tuned CogVideo model across all three enhancement tasks. 
Figure~\ref{fig:qualitative_all} shows the generated progression sequences for super-resolution, deblurring, and low-light enhancement, respectively. Each row represents a single task, and frames 1, 3, 5, 7, and 9 illustrate 
the restoration trajectory over time.

Across all tasks, the qualitative results confirm that the fine-tuned CogVideo successfully learns a temporally consistent enhancement process. For super-resolution, edges and textures become increasingly sharper across frames. For deblurring, motion-induced artifacts are gradually removed while maintaining realistic contours. In the low-light case, brightness and contrast improve steadily with minimal color distortion. Overall, the visual results illustrate that the model not only restores perceptual quality but also encodes the restoration process itself as a smooth temporal evolution, aligning well with our quantitative observations in Section~3.1.

\begin{figure*}[t]
	\centering
	\setlength{\tabcolsep}{2pt}
	\begin{tabular}{ccccc}
		\includegraphics[width=0.19\textwidth]{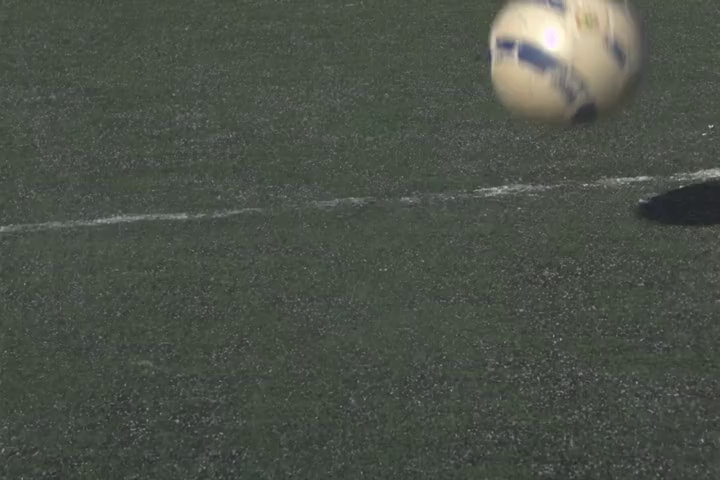} &
		\includegraphics[width=0.19\textwidth]{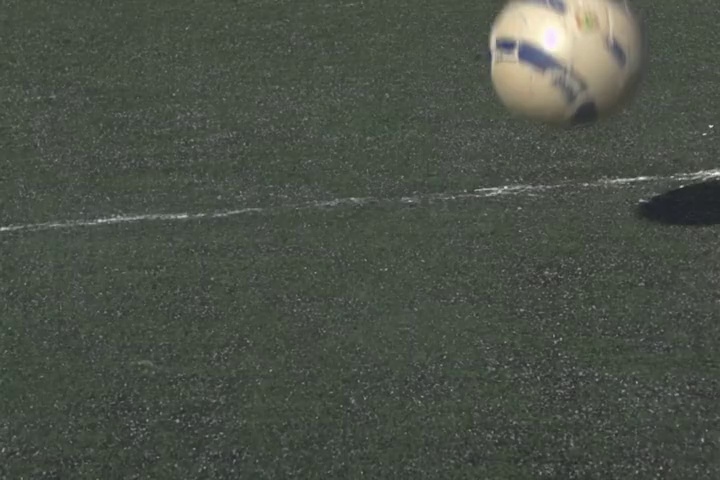} &
		\includegraphics[width=0.19\textwidth]{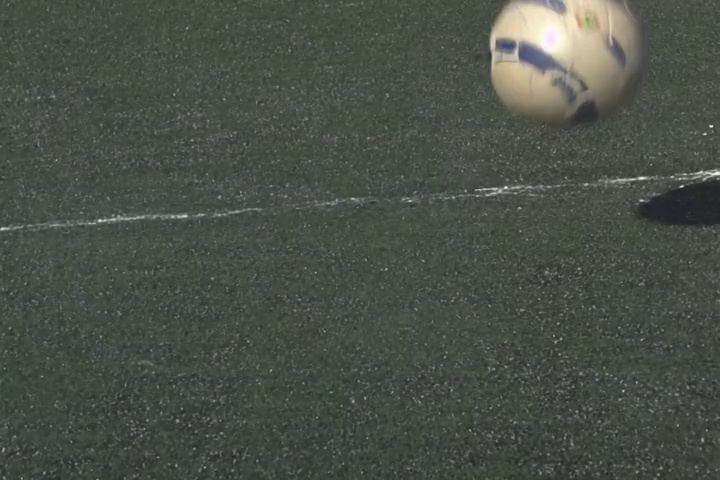} &
		\includegraphics[width=0.19\textwidth]{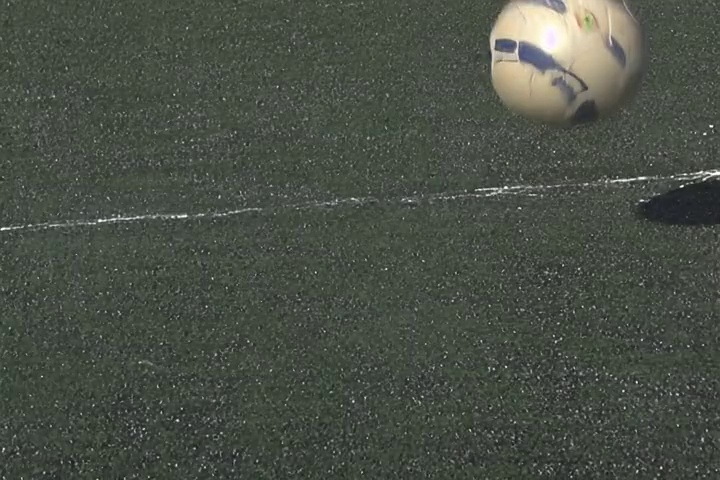} &
		\includegraphics[width=0.19\textwidth]{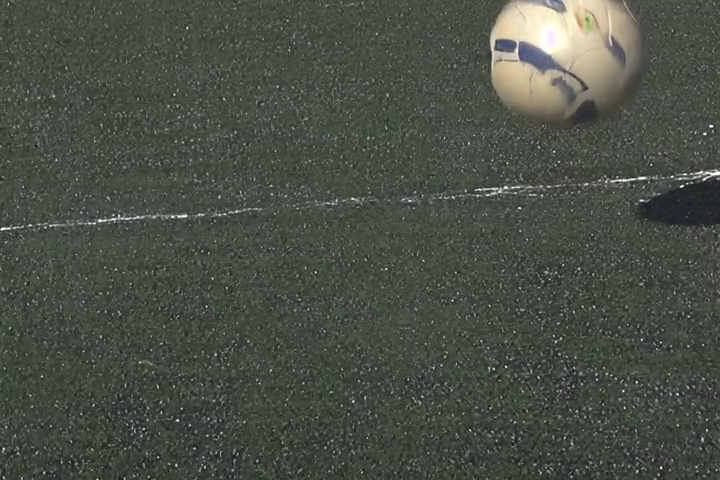} \\
		\small (Frame 1) & \small (Frame 3) & \small (Frame 5) & \small (Frame 7) & \small (Frame 9)
	\end{tabular}
	\caption{
		Qualitative example from the \textbf{ReLoBlur} dataset focusing on a moving football scene. The fine-tuned CogVideo progressively restores the sharp structure and texture of the football while reducing surrounding motion streaks and background smear. From frame~1 to frame~9, the football becomes increasingly clear and well-defined, demonstrating the model’s ability to recover localized high-frequency motion details and maintain temporal consistency without over-sharpening artifacts.
	}
	\label{fig:real_reloblur}
\end{figure*}

\subsection{Evaluation on the Real-World Dataset}

To evaluate the generalization capability of our fine-tuned CogVideo model beyond synthetic training conditions, 
we conduct a zero-shot test on the \textbf{ReLoBlur} dataset, a real-world benchmark that captures natural motion blur across diverse lighting and motion scenarios. No additional fine-tuning or adaptation is performed for this experiment.

As shown in Table~\ref{tab:real_lpips}, the LPIPS metric decreases during the early frames and stabilizes after frame~5, indicating that the model progressively enhances perceptual similarity before convergence. Despite being trained solely on synthetic video progressions, CogVideo generalizes well to real-world motion blur, capturing the temporal evolution from heavily blurred to sharp regions.

Overall, the quantitative and qualitative results on ReLoBlur demonstrate that the proposed fine-tuning strategy enables CogVideo to learn a robust restoration prior that transfers effectively from synthetic to real-world degradations. The ability to model restoration as a temporal generation process appears to promote generalization across diverse motion patterns and lighting conditions.

\section{Conclusion}
\label{sec:conclusion}

We presented a novel approach to image restoration using \textbf{CogVideo} fine-tuned for progressive enhancement. By framing restoration as a temporal generation process, the model learns to transform degraded inputs into gradually improved sequences, covering super-resolution, deblurring, and low-light enhancement. Our results show consistent improvement in perceptual metrics and clear restoration trajectories, with strong generalization to real-world motion blur in the \textbf{ReLoBlur} dataset. This study highlights the potential of text-conditioned video generation as an interpretable and extensible paradigm for unified visual restoration tasks.

\bibliographystyle{IEEEbib}
\bibliography{test}

@article{Liu2023VisualIT,
	title={Visual instruction tuning},
	author={Liu, Haotian and Li, Chunyuan and Wu, Qingyang and Lee, Yong Jae},
	journal={Advances in neural information processing systems},
	volume={36},
	pages={34892--34916},
	year={2023}
}

@article{yang2024cogvideox,
	title={Cogvideox: Text-to-video diffusion models with an expert transformer},
	author={Yang, Zhuoyi and Teng, Jiayan and Zheng, Wendi and Ding, Ming and Huang, Shiyu and Xu, Jiazheng and Yang, Yuanming and Hong, Wenyi and Zhang, Xiaohan and Feng, Guanyu and others},
	journal={arXiv preprint arXiv:2408.06072},
	year={2024}
}

@inproceedings{li2023real,
	title={Real-world deep local motion deblurring},
	author={Li, Haoying and Zhang, Ziran and Jiang, Tingting and Luo, Peng and Feng, Huajun and Xu, Zhihai},
	booktitle={Proceedings of the AAAI Conference on Artificial Intelligence},
	volume={37},
	number={1},
	pages={1314--1322},
	year={2023}
}

@InProceedings{Timofte_2017_CVPR_Workshops,
	author = {Timofte, Radu and Agustsson, Eirikur and Van Gool, Luc and Yang, Ming-Hsuan and Zhang, Lei and Lim, Bee and others},
	title = {NTIRE 2017 Challenge on Single Image Super-Resolution: Methods and Results},
	booktitle = {The IEEE Conference on Computer Vision and Pattern Recognition (CVPR) Workshops},
	month = {July},
	year = {2017}
}

@inproceedings{zamir2022restormer,
	title={Restormer: Efficient transformer for high-resolution image restoration},
	author={Zamir, Syed Waqas and Arora, Aditya and Khan, Salman and Hayat, Munawar and Khan, Fahad Shahbaz and Yang, Ming-Hsuan},
	booktitle={Proceedings of the IEEE/CVF conference on computer vision and pattern recognition},
	pages={5728--5739},
	year={2022}
}

@inproceedings{kim2016accurate,
	title={Accurate image super-resolution using very deep convolutional networks},
	author={Kim, Jiwon and Lee, Jung Kwon and Lee, Kyoung Mu},
	booktitle={Proceedings of the IEEE conference on computer vision and pattern recognition},
	pages={1646--1654},
	year={2016}
}

@inproceedings{tai2017memnet,
	title={Memnet: A persistent memory network for image restoration},
	author={Tai, Ying and Yang, Jian and Liu, Xiaoming and Xu, Chunyan},
	booktitle={Proceedings of the IEEE international conference on computer vision},
	pages={4539--4547},
	year={2017}
}

@article{liu2018non,
	title={Non-local recurrent network for image restoration},
	author={Liu, Ding and Wen, Bihan and Fan, Yuchen and Loy, Chen Change and Huang, Thomas S},
	journal={Advances in neural information processing systems},
	volume={31},
	year={2018}
}

@inproceedings{lefkimmiatis2018universal,
	title={Universal denoising networks: a novel CNN architecture for image denoising},
	author={Lefkimmiatis, Stamatios},
	booktitle={Proceedings of the IEEE conference on computer vision and pattern recognition},
	pages={3204--3213},
	year={2018}
}

@article{radford2018improving,
	title={Improving language understanding by generative pre-training},
	author={Radford, Alec and Narasimhan, Karthik and Salimans, Tim and Sutskever, Ilya and others},
	year={2018},
	publisher={San Francisco, CA, USA}
}

@article{radford2019language,
	title={Language models are unsupervised multitask learners},
	author={Radford, Alec and Wu, Jeffrey and Child, Rewon and Luan, David and Amodei, Dario and Sutskever, Ilya and others},
	journal={OpenAI blog},
	volume={1},
	number={8},
	pages={9},
	year={2019}
}

@article{brown2020language,
	title={Language models are few-shot learners},
	author={Brown, Tom and Mann, Benjamin and Ryder, Nick and Subbiah, Melanie and Kaplan, Jared D and Dhariwal, Prafulla and Neelakantan, Arvind and Shyam, Pranav and Sastry, Girish and Askell, Amanda and others},
	journal={Advances in neural information processing systems},
	volume={33},
	pages={1877--1901},
	year={2020}
}

@article{achiam2023gpt,
	title={Gpt-4 technical report},
	author={Achiam, Josh and Adler, Steven and Agarwal, Sandhini and Ahmad, Lama and Akkaya, Ilge and Aleman, Florencia Leoni and Almeida, Diogo and Altenschmidt, Janko and Altman, Sam and Anadkat, Shyamal and others},
	journal={arXiv preprint arXiv:2303.08774},
	year={2023}
}

@inproceedings{yuan2025identity,
	title={Identity-preserving text-to-video generation by frequency decomposition},
	author={Yuan, Shenghai and Huang, Jinfa and He, Xianyi and Ge, Yunyang and Shi, Yujun and Chen, Liuhan and Luo, Jiebo and Yuan, Li},
	booktitle={Proceedings of the Computer Vision and Pattern Recognition Conference},
	pages={12978--12988},
	year={2025}
}

@article{ruan2024enhancing,
	title={Enhancing motion in text-to-video generation with decomposed encoding and conditioning},
	author={Ruan, Penghui and Wang, Pichao and Saxena, Divya and Cao, Jiannong and Shi, Yuhui},
	journal={Advances in Neural Information Processing Systems},
	volume={37},
	pages={70101--70129},
	year={2024}
}

@inproceedings{lee2024grid,
	title={Grid diffusion models for text-to-video generation},
	author={Lee, Taegyeong and Kwon, Soyeong and Kim, Taehwan},
	booktitle={Proceedings of the IEEE/CVF Conference on Computer Vision and Pattern Recognition},
	pages={8734--8743},
	year={2024}
}

@article{yuan2025magictime,
	title={Magictime: Time-lapse video generation models as metamorphic simulators},
	author={Yuan, Shenghai and Huang, Jinfa and Shi, Yujun and Xu, Yongqi and Zhu, Ruijie and Lin, Bin and Cheng, Xinhua and Yuan, Li and Luo, Jiebo},
	journal={IEEE Transactions on Pattern Analysis and Machine Intelligence},
	year={2025},
	publisher={IEEE}
}

@inproceedings{chen2024videocrafter2,
	title={Videocrafter2: Overcoming data limitations for high-quality video diffusion models},
	author={Chen, Haoxin and Zhang, Yong and Cun, Xiaodong and Xia, Menghan and Wang, Xintao and Weng, Chao and Shan, Ying},
	booktitle={Proceedings of the IEEE/CVF Conference on Computer Vision and Pattern Recognition},
	pages={7310--7320},
	year={2024}
}

@inproceedings{wang2024recipe,
	title={A recipe for scaling up text-to-video generation with text-free videos},
	author={Wang, Xiang and Zhang, Shiwei and Yuan, Hangjie and Qing, Zhiwu and Gong, Biao and Zhang, Yingya and Shen, Yujun and Gao, Changxin and Sang, Nong},
	booktitle={Proceedings of the IEEE/CVF Conference on Computer Vision and Pattern Recognition},
	pages={6572--6582},
	year={2024}
}

@inproceedings{yuan2024instructvideo,
	title={Instructvideo: Instructing video diffusion models with human feedback},
	author={Yuan, Hangjie and Zhang, Shiwei and Wang, Xiang and Wei, Yujie and Feng, Tao and Pan, Yining and Zhang, Yingya and Liu, Ziwei and Albanie, Samuel and Ni, Dong},
	booktitle={Proceedings of the IEEE/CVF Conference on Computer Vision and Pattern Recognition},
	pages={6463--6474},
	year={2024}
}

\end{document}